\documentclass[letterpaper]{article} 
\usepackage{aaai25}  
\usepackage{times}  
\usepackage{helvet}  
\usepackage{courier}  
\usepackage[hyphens]{url}  
\usepackage{graphicx} 
\usepackage{natbib}  
\usepackage{caption} 
\usepackage{algorithm}
\usepackage{algorithmic}
\usepackage{amssymb}
\usepackage{amsmath}
\usepackage{booktabs}
\usepackage{multirow}
\usepackage{newfloat}
\usepackage{listings}
\DeclareCaptionStyle{ruled}{labelfont=normalfont,labelsep=colon,strut=off} 
\floatstyle{ruled}
\newfloat{listing}{tb}{lst}{}
\floatname{listing}{Listing}
\title{IPVTON: Image-based 3D Virtual Try-on with  Image Prompt Adapter}
\author {
    Xiaojing Zhong\textsuperscript{\rm 1,\rm 2},
    Zhonghua Wu\textsuperscript{\rm 3},
    Xiaofeng Yang\textsuperscript{\rm 2},
    Guosheng Lin \textsuperscript{\rm 2}\thanks{Corresponding Authors},
    Qingyao Wu \textsuperscript{\rm 1,\rm 4}\footnotemark[1]
}
\affiliations {
    \textsuperscript{\rm 1}School of Software Engineering, South China University of Technology, China\\
    \textsuperscript{\rm 2}Nanyang Technological University, Singapore\\
    \textsuperscript{\rm 3}SenseTime Research, Singapore\\
    \textsuperscript{\rm 4}Peng Cheng Laboratory, China\\
    vzxj12@gmail.com, gslin@ntu.edu.sg, qyw@scut.edu.cn
}

\usepackage{bibentry}

\begin{document}

\maketitle

\begin{abstract}
Given a pair of images depicting a person and a garment separately, image-based 3D virtual try-on methods aim to reconstruct a 3D human model that realistically portrays the person wearing the desired garment. In this paper, we present IPVTON, a novel image-based 3D virtual try-on framework. IPVTON employs score distillation sampling with image prompts to optimize a hybrid 3D human representation, integrating target garment features into diffusion priors through an image prompt adapter. To avoid interference with non-target areas, we leverage mask-guided image prompt embeddings to focus the image features on the try-on regions. Moreover, we impose geometric constraints on the 3D model with a pseudo silhouette generated by ControlNet, ensuring that the clothed 3D human model retains the shape of the source identity while accurately wearing the target garments. Extensive qualitative and quantitative experiments demonstrate that IPVTON outperforms previous methods in image-based 3D virtual try-on tasks, excelling in both geometry and texture. 
\end{abstract}

%

\section{Introduction}

\begin{figure}[t]
\centering
\includegraphics[width=0.45\textwidth,height=0.3\textheight]{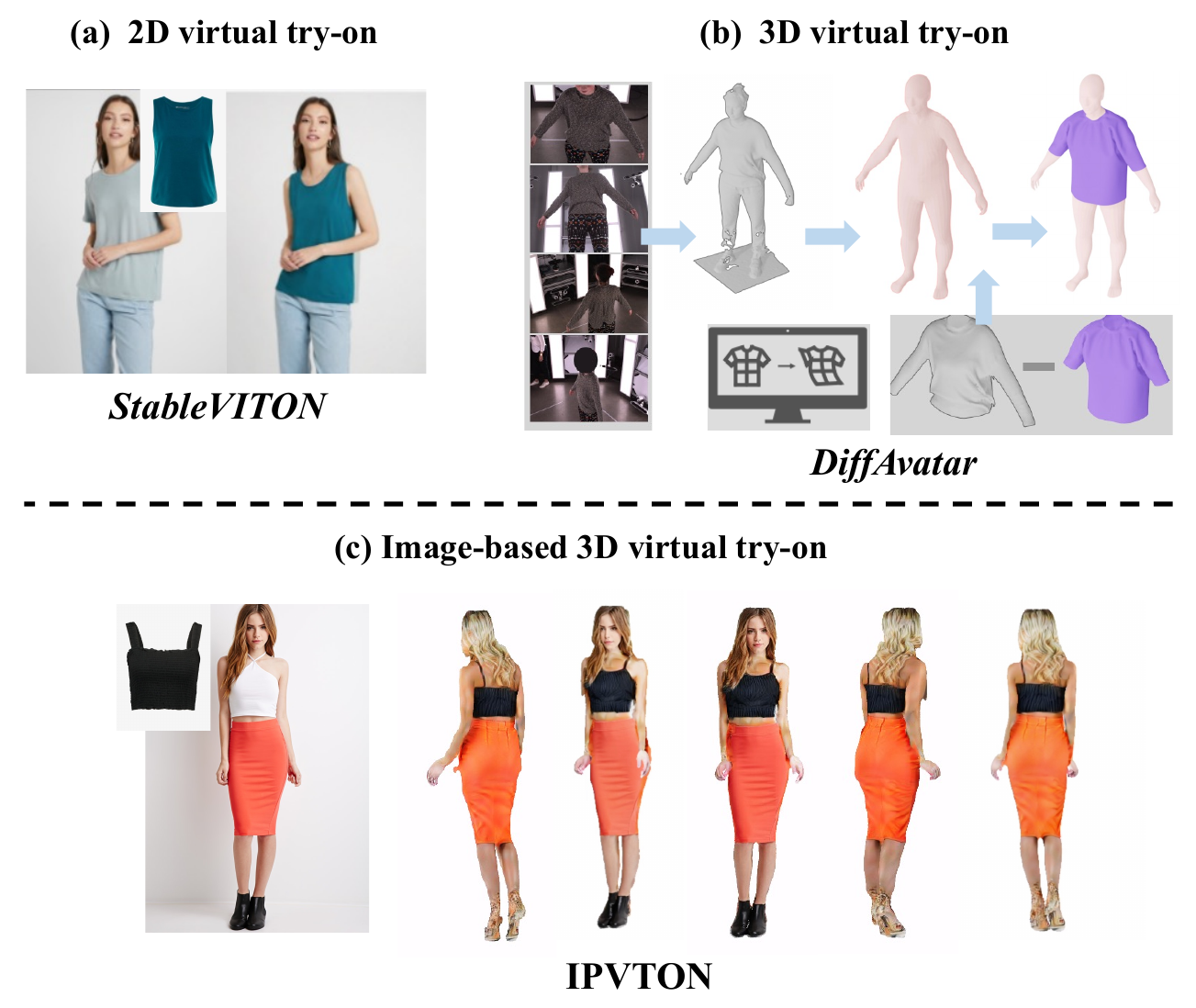}
\caption{Compared to 2D virtual try-on \cite{kim2024stableviton} with its fixed viewpoint and 3D virtual try-on \cite{li2024diffavatar} that require complex processes, IPVTON can generate 3D try-on results from just a human image and a garment image.}
\label{1}
\end{figure}

Human generation has been a prominent task in the AIGC field, with virtual try-on attracting widespread attention due to its significant commercial and entertainment value. Image-based 2D virtual try-on technology, which generates a realistic photo of a person wearing a desired garment by combining the person's image with the garment's image, is valued for its user-friendliness and resource efficiency. However, this method is limited by its reliance on a fixed viewpoint (see Fig. \ref{1} (a)), which poses challenges in real-world applications where users often need to assess the garment from multiple angles. On the other hand, the traditional 3D virtual try-on method provides the advantage of multi-angle views but requires complex processes such as garment-body registration and physics simulations (see Fig. \ref{1} (b)), making it labor-intensive. The challenge of reconstructing accurate 3D models from 2D images, an inherently ill-posed problem, further complicates efforts to integrate image-based and 3D-based virtual try-on techniques.
\par

Owing to the remarkable progress in diffusion models for Text-to-Image (T2I) \cite{ho2020denoising,sohl2015deep,song2019generative}, the field of 3D content generation has seen significant advancements. Recent works \cite{chen2023fantasia3d,wang2024prolificdreamer,qian2023magic123,zhong2025deco} leverage 2D generative priors from pre-trained T2I models (e.g., StableDiffusion (SD)) combined with the Score Distillation Sampling (SDS) loss \cite{poole2022dreamfusion} to optimize 3D representations, resulting in high-quality 3D objects. Despite the success in synthesizing images with specific concepts \cite{ruiz2023dreambooth,kumari2023multi}, extending these techniques to customized 3D object generation remains challenging. For instance, incorporating personalized modules such as LoRAs \cite{hu2021lora} into the SD model diminishes its ability to generate consistent multi-view images \cite{xie2024dreamvton}. Additionally, fine-tuning with only a few images struggles to capture the complex features of garments necessary for 3D virtual try-on. 

 \begin{figure*}[t]
\centering
\includegraphics[width=0.9\textwidth,height=0.4\textheight]{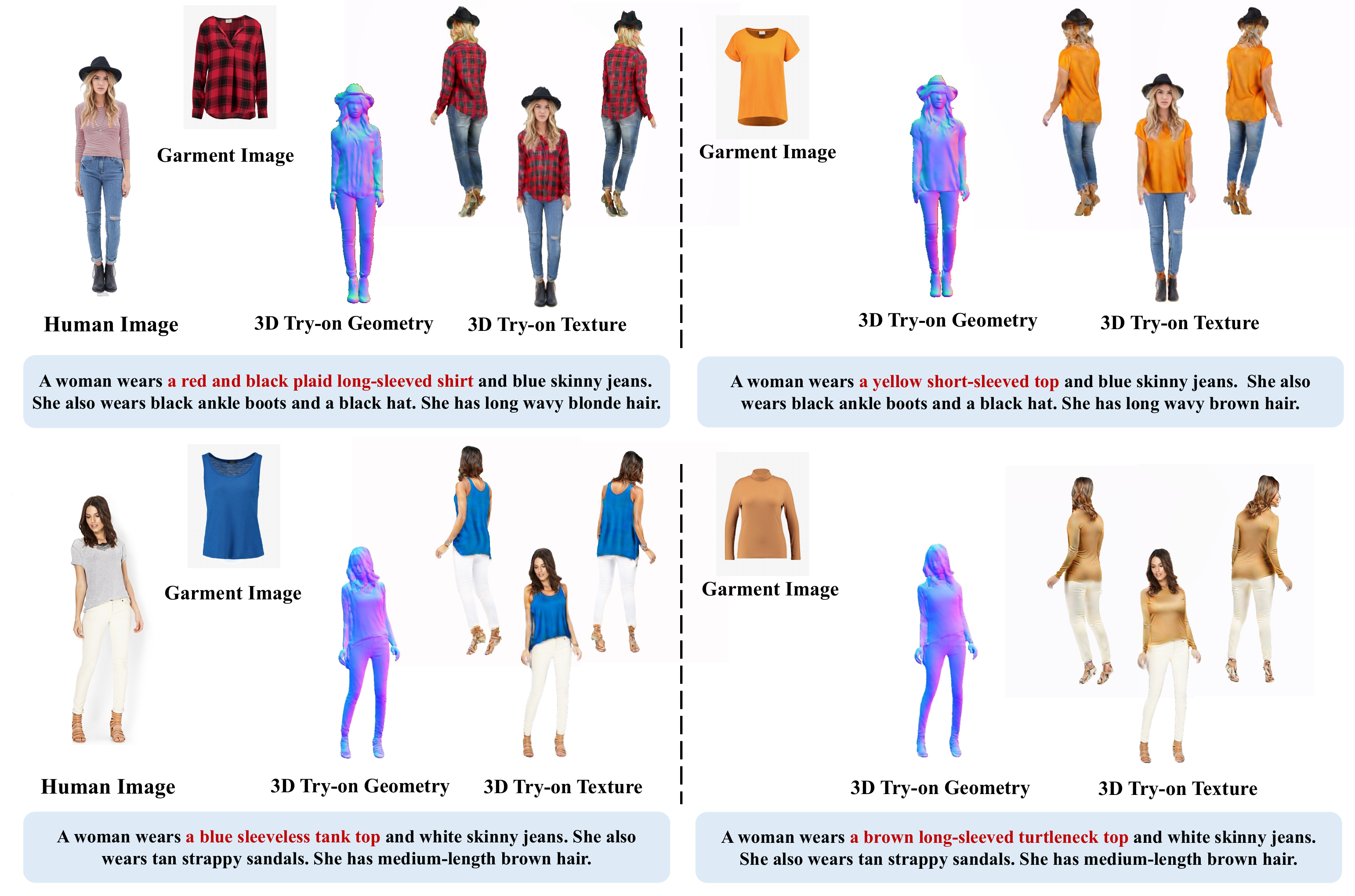}
\caption{\textbf{3D Try-on results.} Given a human image, a garment image and a text prompt, IPVTON can generate realistic 3D human models with the desired garment shapes and textures while preserving the source identity.}
\label{2}
\end{figure*}

Image prompt adapter (IP-Adapter) \cite{ye2023ip} introduces a cross-attention layer for image prompts in diffusion models, enabling controllable generation based on provided images. In this paper, we propose IPVTON, a data-efficient image-based 3D virtual try-on framework that integrates an image prompt adapter with customized diffusion models to optimize a hybrid 3D human model using SDS loss. Since IP-Adapter is compatible with existing diffusion models, it eliminates the need for additional parameter fine-tuning with limited-viewpoint images, preserving multi-view generative priors for consistent 3D model generation. Moreover, combining textual and visual prompts effectively encapsulates the high-level semantics of garments. Specifically, we adopt a two-stage 3D generation framework that independently optimizes the geometry and texture of a hybrid 3D human model initialized with SMPL-X \cite{pavlakos2019expressive}, using an image prompt encoder to extract features from the target garment image and its corresponding normal map to guide the respective optimizations. Unlike using a reference image to influence the entire output \cite{zeng2023ipdreamer,ran2024x}, virtual try-on requires preserving the non-try-on regions of the source human image during optimization. To address this problem, we employ mask-guided image prompt embeddings to focus the image prompt features on the targeted region, reducing unintended effects on surrounding areas. Furthermore, while mask guidance mitigates interference, it may limit the effectiveness of image prompts in guiding geometry generation. To overcome this problem, we introduce a Pseudo Silhouette Loss (PSL) to ensure the generated 3D human conforms to the desired garment shapes. Fig. \ref{2} illustrates the 3D try-on results generated from a human image and an in-shop garment image. Overall, our contributions are summarized as follows:

\begin{itemize}
    \item We design a data-efficient image-based 3D virtual try-on framework that generates 3D human models seamlessly wearing the desired garments, which can be observed from any viewpoint. 
    
    \item We combine score distillation sampling with image prompts to optimize a hybrid 3D human representation, using an image prompt adapter to integrate garment features into the diffusion prior. We leverage mask-guided image prompt embeddings to focus the image features on the masked region, preserving the source identity in non-try-on areas.
    
    \item To ensure the generated model accurately reflects the desired garment shapes, we propose a pseudo silhouette loss to optimize the 3D human geometry.

\end{itemize}


\section{Related Work}
\subsubsection{2D and 3D Virtual Try-on.} Image-based 2D virtual try-on aims to fit an in-shop garment onto a clothed human in an image. Traditional methods primarily rely on Generative Adversarial Networks (GANs) \cite{goodfellow2020generative,zhong2023sara,wu2020exploring,shi2021remember}, where the garment is first deformed to align with the person's pose, followed by a generator that blends the deformed garment with the person’s image \cite{zhong2021mv,wu2019m2e,choi2021viton,ge2021parser}. Building on the advancements of diffusion models in image editing, virtual try-on research has increasingly focused on their application, leveraging pre-trained diffusion models to blend garments with human appearances \cite{kim2024stableviton,choi2024improving,zhu2023tryondiffusion}. Despite the success of 2D virtual try-on methods, they struggle to generate multi-view try-on results, which are crucial for real-world applications. 

With the increasing demand for 3D virtual try-on, \cite{bhatnagar2019multi,mir2020learning,patel2020tailornet,zhong2023di} represent garments layered over the SMPL model \cite{loper2023smpl,pang2024towards}. M3D-VTON \cite{zhao2021m3d} constructs a 3D clothed human by predicting dual depth maps for a person’s image and applies these depth values to the results of 2D virtual try-on. To leverage the powerful generative prior of diffusion models, DreamVTON \cite{xie2024dreamvton} combines SDS loss with LoRAs \cite{hu2021lora} to generate 3D humans with customized identities and clothing. However, the need to fine-tune the LoRA layers for each pair of samples incurs a time cost. Efficiently integrating desired garment features into a diffusion model remains a challenge.

\subsubsection{Text-guided 3D Human Generation.} Avatar-CLIP \cite{hong2022avatarclip} initializes the geometry of 3D human using a shape VAE network and refines geometry and texture with CLIP loss \cite{radford2021learning}. Dreamwaltz \cite{huang2024dreamwaltz} improves SDS loss by incorporating 3D-aware skeleton conditioning, while Humannorm \cite{huang2024humannorm} and AvatarVerse \cite{zhang2024avatarverse} utilize the hybrid 3D representation DMTet \cite{shen2021deep} combined with structural condition maps to achieve more detailed and realistic geometry. TADA \cite{liao2024tada} enhances the upsampled SMPL-X model by adding a displacement layer and texture map. TeCH \cite{huang2024tech} combines SDS loss with DreamBooth. However, while the geometry and texture of the generated 3D human can be altered by modifying the text prompt, the results often deviate from the provided image. 

\subsubsection{Customizing Diffusion Models.}

DreamBooth \cite{ruiz2023dreambooth} fine-tunes the network on a small set of subject-specific images, enabling the customization of diffusion models to closely match the style or subject of the provided images. LoRA \cite{hu2021lora} reduces trainable parameters by learning rank-decomposition matrices, enabling efficient fine-tuning of pre-trained diffusion models with specific concepts. Custom Diffusion \cite{kim2024stableviton} fine-tunes a small subset of weights in the cross-attention layers, focusing on the key and value mappings from text to latent features. IP-Adapter \cite{ye2023ip} proposes decomposing the cross-attention layers for text and image features, allowing an image prompt adapter to incorporate additional image styles. \cite{choi2024improving} first customize a diffusion model with IP-Adapter for 2D virtual try-on. IPDreamer \cite{zeng2023ipdreamer} combines SDS loss with IP-Adapter, enabling the customization of 3D models. However, since it is designed for general 3D objects, applying it directly to humans yields coarse results due to the complexity of human topology.

\section{Preliminaries}

\textbf{Latent Diffusion Model (LDM)} performs diffusion in a lower-dimensional latent space for decreasing computing cost. Specifically, LDM employs an autoencoder to encode an input image $x$ into a latent code $z=\mathcal{E}(x)$ and decode $z$ to $x=\mathcal{D}(z)$. During the forward stage, the initial latent code $z_0$ is gradually perturbed by adding Gaussian noise $\epsilon$ over the time step $t$ to match the Gaussian distribution: $z_t \sim \mathcal{N}(0,I)$. In the reverse stage, a noise predictor $\epsilon_{\phi}$ based on a U-Net structure \cite{ronneberger2015u} and parameterized by $\phi$ is trained to predict the noise added at each corresponding step of the forward process. The training uses the following loss function:  

\begin{equation}
    \min _{\phi} \mathbb{E}_{\mathbf{z}_0, \epsilon \sim \mathcal{N}(0, I), t}\left\|\epsilon_{\phi}\left(z_t;y,t\right)-\epsilon\right\|_2^2,
\end{equation}
where $y$ represents a conditional text prompt and $\epsilon$ denotes the added random noise.

\textbf{Score Distillation Sampling (SDS)} is proposed to optimize a 3D representation parameterized by $\eta$ using differentiable rendering, ensuring that the rendered 2D images conform to the diffusion prior. Given a random camera pose, the differentiable rendering function $\mathbf{g}$ generates the rendered image $I$ via $I=\mathbf{g}(\eta)$. $\eta$ is optimized for 3D consistency by computing the gradient of $\mathcal{L}_{\mathrm{SDS}}$ with respect to $z$, which is encoded from the rendered image $I$:

\begin{equation}\label{eq.1}
    \nabla_{\eta} \mathcal{L}_{\mathrm{SDS}}(\phi, z)=\mathbb{E}_{t, {\epsilon}}\left[w(t)\left(\hat{\epsilon}_\phi\left(z_t ; y, t\right)-\epsilon\right) \frac{\partial z}{\partial \eta}\right],
\end{equation}
where $w(t)$ is a time-dependent weighting function that varies with $t$ and $z_t$ is the noised latent vector. Compared to $\epsilon_\phi$, $\hat{\epsilon}_\phi$ incorporates classifier-free guidance \cite{ho2022classifier} to align the diffusion process with the target prompt. 

\begin{figure*}[t]
\centering
\includegraphics[width=1.0\textwidth,height=0.4\textheight]{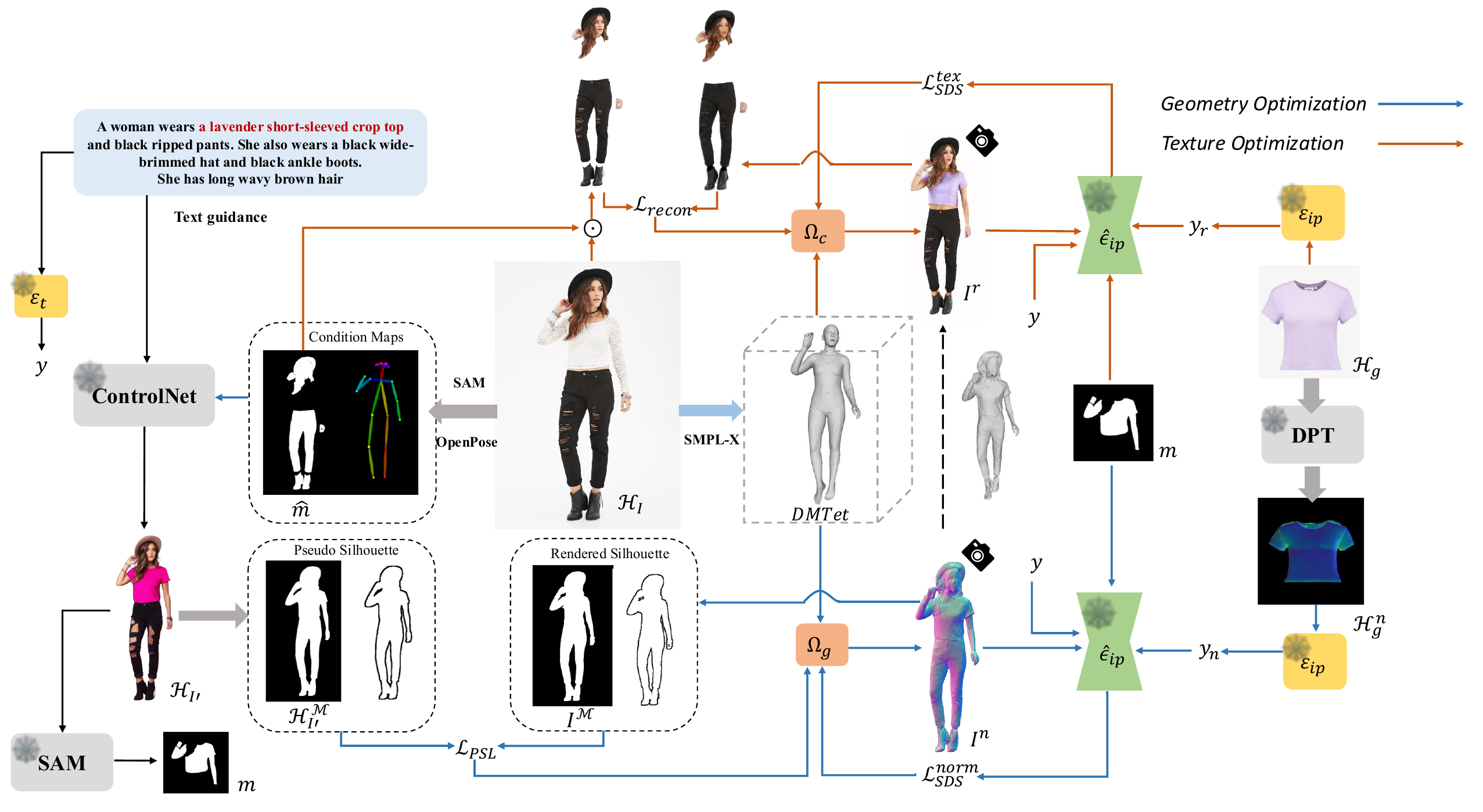}
\caption{\textbf{Overview of IPVTON.}  Given a human image $\mathcal{H}_I$, we first construct a DMTet-based 3d representation initialized with SMPL-X to model the human, with its geometry and texture generated through $\Omega_g$ and $\Omega_c$, respectively. During geometry optimization, the rendered human normal map $I^n$ is encoded into the diffusion model $\hat{\epsilon}_{ip}$ and, along with $y_n$ and $m$, is used to compute $\mathcal{L}_{SDS}^{norm}$. $y_n$ is the normal image prompt embedding encoded from $\mathcal{H}_g^n$ via $\mathcal{E}_{ip}$, and $m$ is a mask covering the try-on region, derived from $\mathcal{H}_I'$.  During texture optimization, the rendered human image $I^r$ is encoded into $\hat{\epsilon}_{ip}$ and along with $y_r,y$ and $m$, is used to compute $\mathcal{L}_{SDS}^{tex}$. $y$ is the text prompt embedding encoded from the target texts via $\mathcal{E}_{t}$, and $y_r$ is the image prompt embedding encoded from $\mathcal{H}_g$ via $\mathcal{E}_{ip}$. $\odot$ denotes pixel-wise multiplication.}
\label{3}
\end{figure*}

\section{Method}


We first introduce an efficient 3D hybrid representation, initialized with the SMPL-X human body prior \cite{loper2023smpl}, to model the source identity's body shape and pose. Building on this model, we adopt a two-stage, text-guided 3D generation framework that independently optimizes the geometry and texture using SDS loss with mask-guided image prompt embeddings. To ensure the generated 3D human conforms to the desired garment shape, we employ a pseudo silhouette loss to constrain the geometry generation.

\subsection{3D Hybrid Human Representation}

We utilize DMtet \cite{shen2021deep} as our 3D representation because it combines explicit and implicit forms to efficiently model the 3D clothed human and can be easily converted into meshes. Inspired by \cite{huang2024tech}, we create an outer shell $M_{shell}$ of SMPL-X \cite{feng2021collaborative} to form an outer shell tetrahedral grid $(V_{shell}, T_{shell}$), with $V_{shell}$ representing the set of vertices and $T_{shell}$ representing the set of tetrahedrons in the grid. For each vertex $v_i \in V_{shell}$, we train an MLP-based neural network $\Omega_g$, parameterized by $\phi_g$, to predict its Signed Distance Field (SDF) value: $\Omega_g(v_i) = s(v_i;\phi_g)$. We initialize $\Omega_g$ as follows:

\begin{equation}
    \mathcal{L}_{SDF}(\Omega_g)=\sum_{x \in \mathbf{P}}\left\|s\left(x ; \phi_g\right)-\operatorname{SDF}\left(x\right)\right\|_2^2,
\end{equation}
where $\mathbf{P}$ is the set of random sampling points near $M_{shell}$. 
\par
Next, we employ the Marching Tetrahedra (MT) algorithm \cite{doi1991efficient} for iso-surface extraction, resulting in triangular meshes. Additionally, we train another MLP-based neural network $\Omega_c$, parameterized by $\phi_c$, to generate its albedo map. Given a sampled camera pose, we can utilize differentiable rasterization \cite{laine2020modular} to render the human mesh's normal map $I^n$, color map $I^r$, and mask $I^\mathcal{M}$.


\subsection{Geometry Optimization Stage}

To optimize geometry guided by text prompts, \cite{chen2023fantasia3d,wang2024prolificdreamer,huang2024humannorm} encode the rendered normal map, with the resulting encoding serving as input to the diffusion model for calculating the normal SDS loss. However, representing garments with complex shapes remains challenging without prompt engineering, as text prompts typically describe limited garment dimensions (e.g., length, width). To capture the high-level semantics of garments, we utilize IP-Adapter \cite{ye2023ip} to combine textual and image prompts with a decoupled cross-attention mechanism. Specifically, an image prompt adapter $\mathcal{E}_{ip}$ is used to project the image into a sequence of features that are combined with the textual embedding. As shown in Fig. \ref{3}, we extract the image prompt feature of the target garment's normal map, denoted as $y_n=\mathcal{E}_{ip}(\mathcal{H}_g^n)$, where the normal image $\mathcal{H}_g^n$ is obtained from the target garment image $\mathcal{H}_g$ via normal map estimation DPT \cite{ranftl2021vision}. The calculation of normal SDS loss is as follows:

\begin{equation}\label{eq3}
\begin{split}
   & \nabla_{\phi_g} \mathcal{L}_{SDS}^{norm}(\phi', z^n)=
   \\
   & \mathbb{E}_{t, {\epsilon}}\left[w(t)\left(\hat{\epsilon}_{ip}\left(z^n_t ; y, y_n, t\right)-\epsilon\right) \frac{\partial z^n}{\partial \phi_g}\right],
\end{split}
\end{equation}

\noindent where $\hat{\epsilon}_{ip}$ refers to the diffusion model employed in IP-Adapter, with $\phi'$ denoting its parameters, and $z^n$ represents the latent codes encoded from $I^n$.
However, using global image prompt embeddings can unintentionally affect the entire body, including areas that should remain unchanged during geometry optimization. To address this problem, we employ mask-guided image prompt embeddings to focus the image prompt features on the masked region. Given the query features $\textbf{Z}$, the output of the cross-attention module for the image prompt, $\mathbf{Z'}$, is computed as follows:

\begin{equation}\label{eq4}
    \mathbf{Z'}= m\operatorname{Softmax}\left(\frac{\mathbf{QK}_{ip}^{\top}}{\sqrt{d}}\right) \mathbf{V}_{ip},
\end{equation}

\noindent where $\mathbf{Q}=\mathbf{ZW}_q,\mathbf{K}_{ip}=y_n\mathbf{W'}_k$ and $\mathbf{V}_{ip}=y_n\mathbf{W'}_v$ are the query, key, and value matrices for the normal image prompt features $y_n$. $\mathbf{W}_q,\mathbf{W'}_k,\mathbf{W'}_v$ are the projection matrices used for linear transformation. $m$ denotes the partial mask of the pseudo mask $\mathcal{H}^\mathcal{M}_{I’}$ generated by SAM, covering the try-on regions, which will be described later.

Although mask guidance reduces interference, it can restrict the ability of image prompts to effectively steer geometry generation. To enforce geometric constraints on the generated 3D model, we apply a pseudo silhouette loss to shape the contours of the 3D human. Specifically, as shown in Fig. \ref{3}, we use two condition maps to guide ControlNet: a partial segmentation map of $\mathcal{H}_I$, excluding the regions to be transferred, and a pose map of $\mathcal{H}_I$, extracted with OpenPose \cite{cao2017realtime}. This process generates a human image $\mathcal{H}_{I'}$ that combines the body from $\mathcal{H}_I$ with the garment shapes from $\mathcal{H}_g$. We then use SAM to generate the corresponding mask, $\mathcal{H}^\mathcal{M}_{I'}$. The pseudo silhouette loss can be formulated as follows:


\begin{equation}\label{eq1}
\begin{split}
 &\mathcal{L}_{\text {PSL}}=\|\mathcal{H}^\mathcal{M}_{I'}-I^\mathcal{M})\|_2^2 + \\
& \sum_{k \in \operatorname{Edge}(I^\mathcal{M})} \min _{\hat{k} \in \operatorname{Edge}(\mathcal{H}^\mathcal{M}_{I'})}\|k-\hat{k}\|_1.
\end{split}
\end{equation}

It ensures that both the edges and the silhouette mask of $I^\mathcal{M}$ align with those of $\mathcal{H}^\mathcal{M}_{I’}$. Moreover, we can estimate the normal maps of $\mathcal{H}_I$ and $\mathcal{H}_{I'}$ using ICON \cite{xiu2022icon}. By combining partial normal maps from $\mathcal{H}_I$ and $\mathcal{H}_{I'}$ based on segmentation maps, we can also obtain a pseudo normal map ground truth $\mathcal{H}_I^{n'}$ to further constrain the geometry:

\begin{equation}\label{eq2}
    \mathcal{L}_{norm}= \| \mathcal{H}_I^{n'}-I^n \|_2^2.
\end{equation}

Note that the camera views used to render $I^\mathcal{M}$ and $I^n$ in Eq. \ref{eq1} and Eq. \ref{eq2} are specifically the front and back views. The overall geometry loss functions are calculated as follows: 

\begin{equation}\label{eq5}
\begin{aligned}
\mathcal{L}_{geo} & = \lambda_{PSL}\mathcal{L}_{PSL} + \lambda_{norm}\mathcal{L}_{norm} \\
&+ \lambda_{SDS}^{norm}\mathcal{L}_{SDS}^{norm} + \lambda_{lap}\mathcal{L}_{lap},
\end{aligned}
\end{equation}

\noindent where $\lambda_{\{PSL,norm,SDS(norm),lap\}}$ denotes the weights used to balance the geometry losses, and $\mathcal{L}_{lap}$ represents the Laplacian smoothing term \cite{ando2006learning}, applied for surface regularization. 

\subsection{Texture Optimization Stage}

Despite the guidance provided by text prompts, accurately capturing the target garment's texture remains challenging, as text descriptions often fail to convey its brightness and saturation. We extract the image prompt embedding of the target garment as $y_r=\mathcal{E}_{ip}(I^g)$. The texture SDS loss $\mathcal{L}_{SDS}^{tex}$ is obtained as follows: 


\begin{equation}\label{eq6}
\begin{split}
  &  \nabla_{\phi_c} \mathcal{L}_{SDS}^{tex}(\phi', z^r)=
    \\
    &\mathbb{E}_{t, {\epsilon}}\left[w(t)\left(\hat{\epsilon}_{ip}\left(z^r_t ; y, y_r, t\right)-\epsilon\right) \frac{\partial z^r}{\partial \phi_c}\right],
\end{split}
\end{equation}

\noindent where $z^r$ represents the latent codes encoded from $I^r$. 

Similar to the geometry optimization, we apply mask $m$ to the image prompt features $y_r$ to concentrate the garment texture on the target region. Given the query features $\hat{\mathbf{Z}}$, the output of cross-attention for $y_r$ is denoted as $\mathbf{Z''}$:

\begin{equation}\label{eq7}
    \mathbf{Z''}= m\operatorname{Softmax}\left(\frac{\mathbf{Q'K}_{ip'}^{\top}}{\sqrt{d}}\right) \mathbf{V}_{ip'},
\end{equation}

\noindent where $\mathbf{Q'}=\mathbf{\hat{Z}W}_q,\mathbf{K}_{ip'}=y_r\mathbf{W'}_k$ and $\mathbf{V}_{ip'}=y_r\mathbf{W'}_v$ represent the query, key, and value matrices of the cross-attention module for image prompt features $y_r$.

To retain the appearance of the source human image in regions unaffected by the garment transfer, we employ $\hat{m}$ to constrain the local texture as follows: 

\begin{equation}\label{eq8}
    \mathcal{L}_{recon}= \| \hat{m}(\mathcal{H}_I-I^r) \|_2^2,
\end{equation}

\noindent where $\hat{m}$ represents the mask extracted from $\mathcal{H}_I$ that excludes the regions to be transferred. The overall texture loss functions are calculated as follows: 

\begin{equation}\label{eq9}
\begin{aligned}
\mathcal{L}_{tex} = \lambda_{SDS}^{tex}\mathcal{L}_{SDS}^{tex} + \lambda_{recon}\mathcal{L}_{recon},
\end{aligned}
\end{equation}
where $\lambda_{\{recon,SDS(tex)\}}$ denotes the weights used to balance the texture losses.

\begin{figure*}[ht]
\centering
\includegraphics[width=0.95\textwidth,height=0.63\textheight]{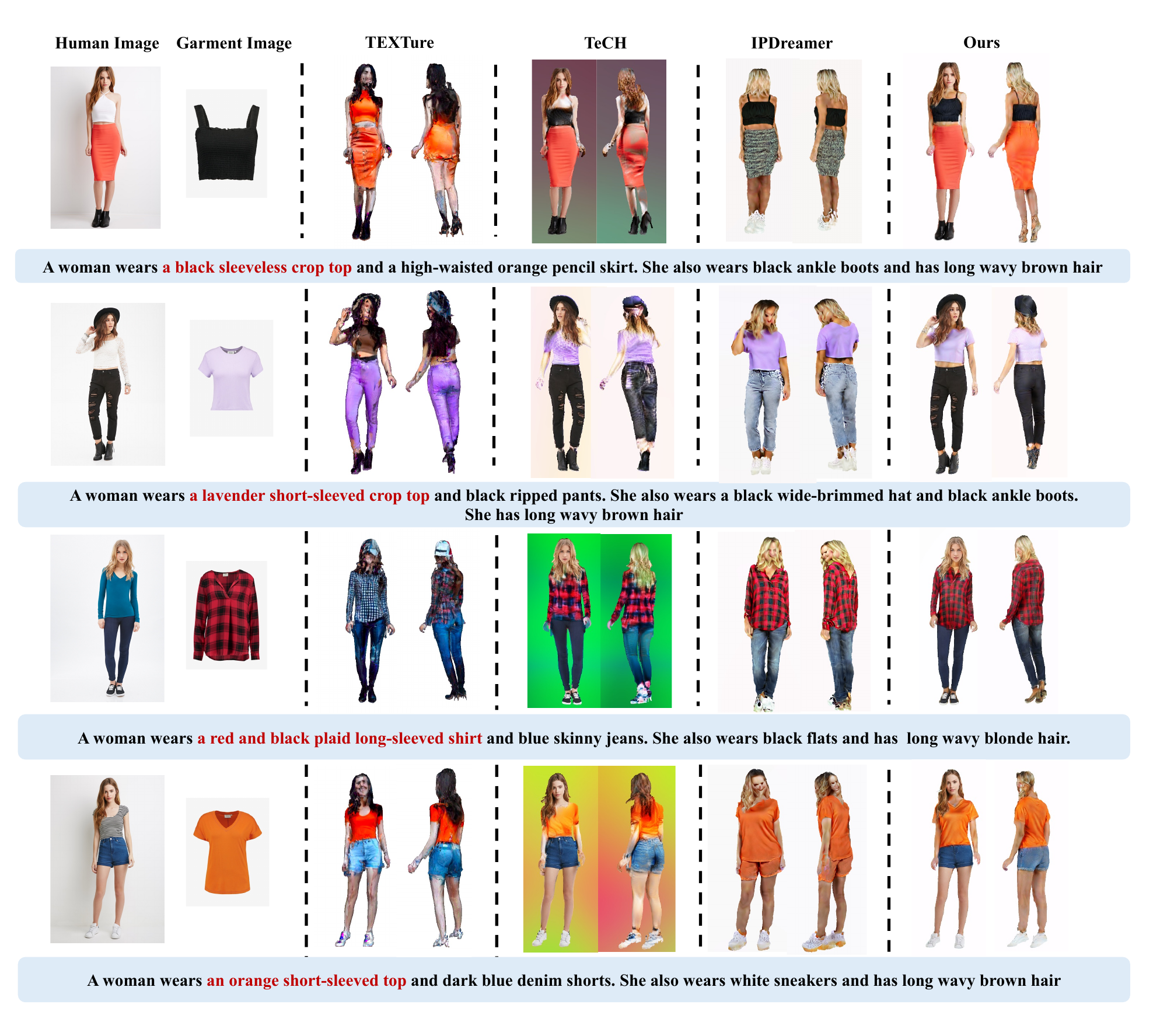}
\caption{\textbf{Qualitative comparisons.} Our IPVTON is able to generate realistic 3D try-on results with high-quality textures, viewable from multiple angles.}
\label{4}
\end{figure*}

\section{Experiment}

\subsubsection{Implementation Details.} We train both $\Omega_g$ and $\Omega_c$ for 100 iterations with one GeForce RTX 2080 Ti GPU. During the geometry optimization stage, we set $\lambda_{PSL}$, $\lambda_{norm}$, and $\lambda_{lap}$ to 10,000, and set $\lambda_{SDS}^{norm}$ to 1. During the texture optimization stage, we set $\lambda_{recon}$ and $\lambda_{SDS}^{tex}$ to 10,000 and 1, respectively.

\subsubsection{Datasets.} Our method does not require paired human and clothing images for training. We select 12 full-body, front-facing human images of different individuals from the DeepFashion dataset \cite{shen2021deep}. For each human image, we choose two garment templates from the VITON-HD dataset \cite{choi2021viton}, covering various types such as tank tops, short sleeves, and long sleeves. Descriptive text prompts for both the human and garment images are generated using ChatGPT-4o. More details are provided in the supplementary material. 

\subsubsection{Baselines.} To demonstrate the effectiveness of our proposed IPVTON, we conduct a comparative analysis with the following baseline methods. 1) $\textit{TEXTure}$ \cite{richardson2023texture} generates and edits the texture of 3D objects based on text prompts. 2) $\textit{TeCH}$ \cite{huang2024tech} leverages SDS loss with fine-tuned DreamBooth \cite{ruiz2023dreambooth} for text-guided 3D human reconstruction. 3) $\textit{IPDreamer}$ \cite{zeng2023ipdreamer} utilizes IP-Adapter to control both the geometry and appearance of 3D objects. Since TEXTure can generate textures but not geometry, we downsample the 3D human model generated by TeCH for faster UV unwrapping with an atlas, using it as the target for texture generation. To ensure a fair comparison, we apply the mask $\hat{m}$ to the reconstruction loss used in TeCH, so that only the texture of the try-on regions is affected.

\begin{figure*}[hpt]
\centering
\includegraphics[width=0.9\textwidth,height=0.25\textheight]{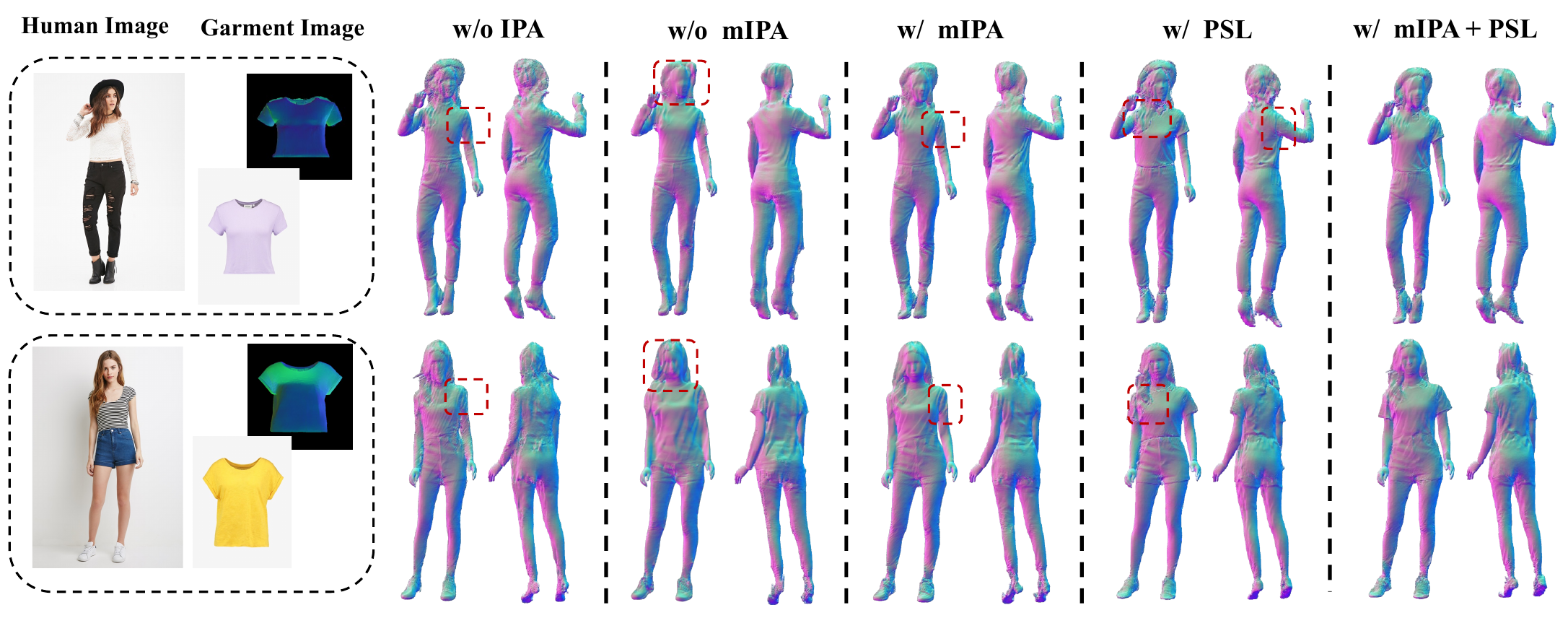}
\caption{\textbf{Ablation study for geometry optimization.} $\texttt{`}$mIPA$\texttt{`}$ denotes mask-guided image prompt embeddings.}
\label{5}
\end{figure*}

\subsubsection{Qualitative Comparison.} As shown in Fig. \ref{4}, TEXTure struggles to generate accurate garment textures and correctly position them on the body according to the target prompt. While TeCH can produce try-on results, it faces challenges in reshaping the human body to match the desired garment shapes. IPDreamer, by leveraging IP-Adapter, captures garment features effectively, accurately reflecting the garment's color, length, and style. However, this method, designed for general 3D objects, results in a coarse human appearance and fails to distinguish between the front and back of the person. In contrast, our IPVTON generates realistic 3D try-on results that capture the desired garment shapes and textures while preserving the source identity in non-try-on areas.

\subsubsection{Quantitative Comparison.} We employ CLIP \cite{radford2021learning} metrics to quantitatively evaluate the faithfulness of the generated 3D try-on results to the target text prompts. We select 8 sets of human images with different identities, each paired with a garment template distinct from the clothing worn by the individuals. For each generated 3D model, we render images from six uniformly sampled angles. CLIP scores are computed by comparing the CLIP embeddings of these images with the target text prompt embeddings. To evaluate geometry faithfulness, we remove texture-related words from the target text prompt and prepend $\texttt{`}the\,normal\,map\,of\texttt{`}$. As shown in Tab. \ref{tab1}, our method achieves the best scores for both geometry and texture faithfulness. We also conduct a user study to further evaluate our method. Using 8 sets of 3D try-on results generated by four methods, we invite 15 volunteers to rank each set according to their preferences for geometry and texture. For each set, users are presented with the source human image, the garment image, and the target text prompt. Participants rank the results separately for geometry and texture on a scale from 4 (highest) to 1 (lowest), without repeating scores. The final report presents the average scores across all sets. As shown in Table \ref{tab1}, our method achieves the highest human preference in both geometry and texture.

\begin{table}[t]
    \centering
    \resizebox{0.46\textwidth}{!}{
    \begin{tabular}{lccccc}
        \toprule
        \multirow{2}{*}{Methods} & \multicolumn{2}{c}{CLIP $\uparrow$} & \multicolumn{2}{c}{User $\uparrow$} \\ 
        & Geo-Faith  & Tex-Faith & Geometry & Texture \\
        \midrule 
        TEXTure  & 30.85 & 28.31 & 2.34 & 1.53 \\ 
        TeCH & 31.41 & 32.34 & 2.55 & 2.6 \\ 
        IPDreamer  & 28.53  & 30.49 & 2.28 & 1.86 \\ 
        IPVTON & \textbf{31.77} & \textbf{33.60} & \textbf{2.63} & \textbf{3.52} \\ 
        \bottomrule
    \end{tabular}}
    \caption{Quantitative evaluation of the results obtained from different methods. $\texttt{`}$Geo-Faith$\texttt{`}$ and $\texttt{`}$Tex-Faith$\texttt{`}$ respectively denote the geometry and texture faithfulness. The best scores are highlighted in \textbf{bold}.} \label{tab1}
\end{table}

\subsubsection{Ablation Study.} 
\begin{figure}[t]
\centering
\includegraphics[width=0.45\textwidth,height=0.23\textheight]{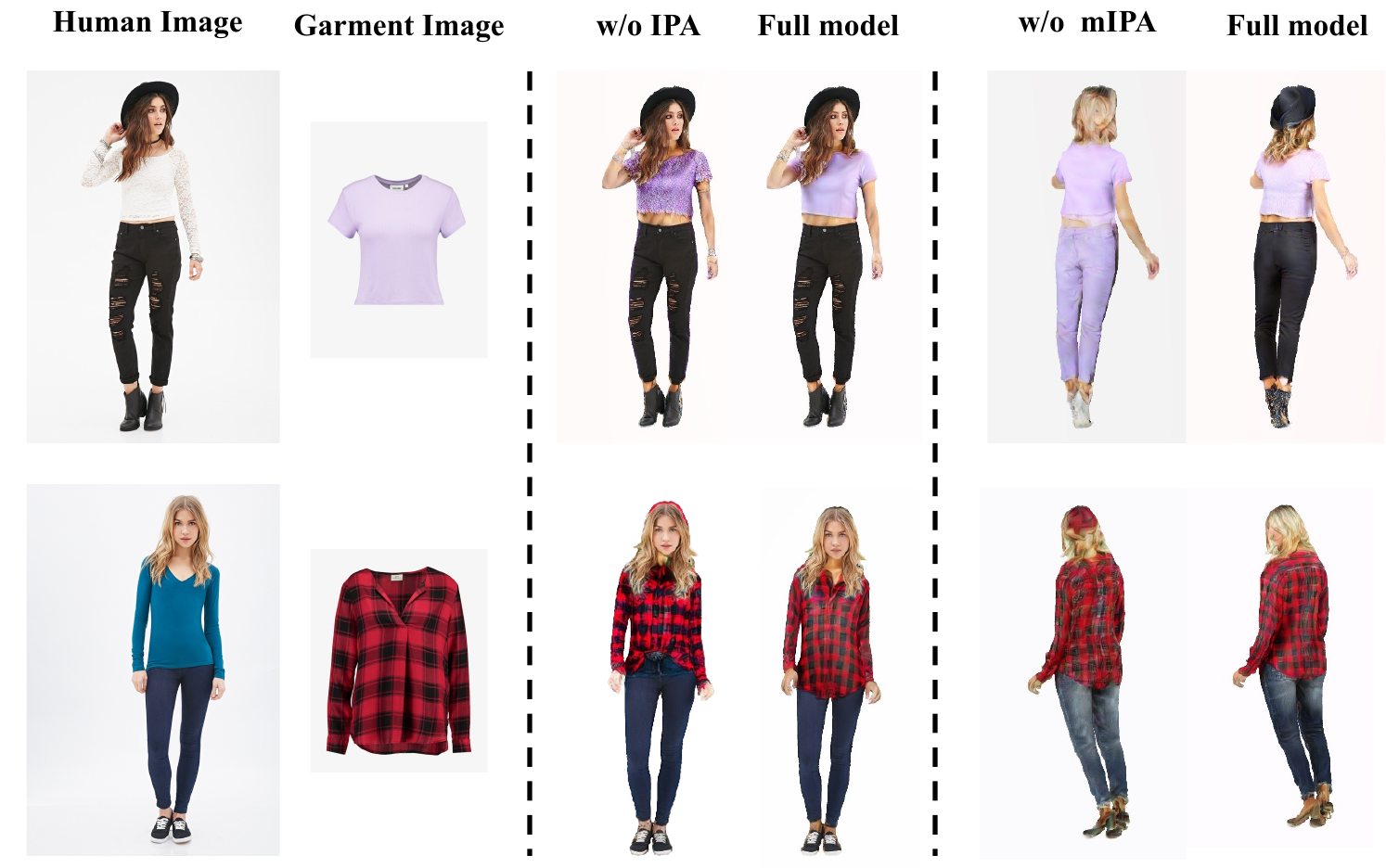}
\caption{\textbf{Ablation study for texture optimization.} $\texttt{`}$mIPA$\texttt{`}$ denotes mask-guided image prompt embeddings.}
\label{6}
\end{figure}

As shown in Fig. \ref{5}, the geometry of the human model generated without using the image prompt adapter retains the geometry of the source human image but fails to reflect the desired garment shapes. In the second column, using global image prompt embeddings allows the body shape to adopt the garment's contours, but this also affects other parts, leading to a blurred face. In comparison, mask-guided image prompt embeddings preserve the source body shapes with sharp details, even though the garment shapes are not fully realized. Note that the results in the third column differ from those in the first because the third column includes garment features like the collar. Relying solely on PSL can cause noisy seams and inaccurate shapes, as seen in the back views of the fourth column, due to potential inaccuracies in the generated pseudo silhouette. Combining mask-guided image prompt embeddings with PSL supervision, IPVTON accurately generates the desired garment contours while maintaining well-defined human body shapes. As shown in Fig. \ref{6}, when texture is generated solely from text prompts, the resulting texture corresponds to the text prompt but deviates from the garment image. For instance, in the first row of the first column, the lavender crop top generated without using the image prompt adapter is slightly darker than the garment image. In the second column, when only the top is meant to be changed, the pants and hair are also affected if image prompt embeddings are used without mask guidance.

\section{Limitation}

Reconstructing an extremely loose target garment may fail, likely due to inherent limitations of the SMPL-X initialization. Additionally, since we use a pre-trained image prompt adapter designed for high-level semantic features, the resulting features may not accurately capture the garment's complex patterns or logos. More details are provided in the supplementary material.

\section{Conclusion}

We propose an image-based 3D virtual try-on framework that optimizes 3D models by integrating garment features via a customized diffusion model with an image prompt adapter. Mask-guided prompt embeddings focus on try-on regions, minimizing interference. A pseudo silhouette loss constrains the 3D geometry, shaping the human form with the desired garment and source identity.

\section{Acknowledgments}
This work was supported by National Natural Science Foundation of China (NSFC) 62272172, Guangdong Basic and Applied Basic Research Foundation 2023A1515012920.
Zhuhai Science and Technology Plan Project(2320004002758). This research is partly supported by the MoE AcRF Tier 2 grant (MOE-T2EP20223-0001). 

\bibliography{aaai25}

\end{document}


\maketitle
\section{Ablation Study}

\begin{figure}[t]
\centering
\setlength{\abovecaptionskip}{0pt}
\includegraphics[width=0.45\textwidth,height=0.26\textheight]{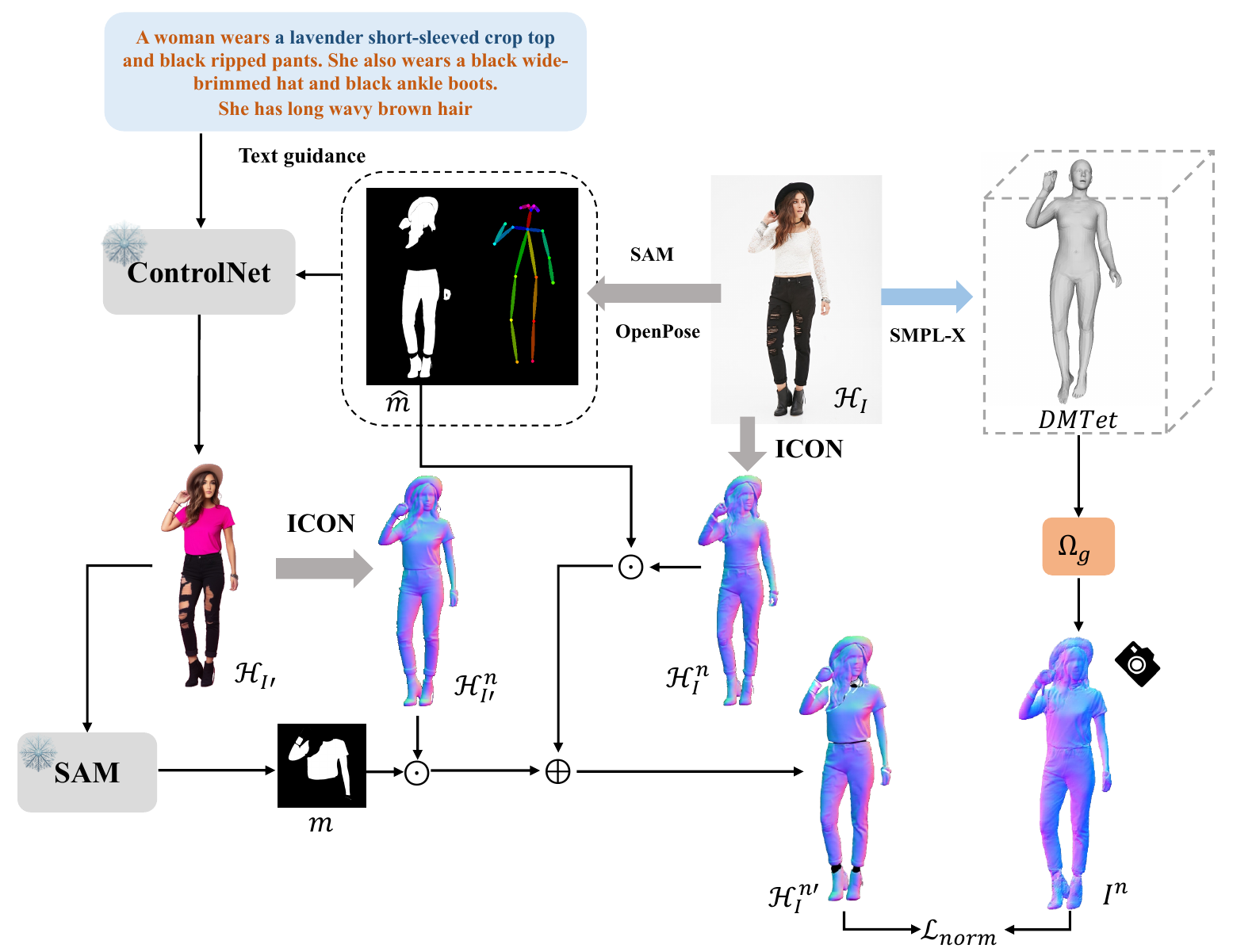}
\caption{\textbf{The calculation of $\mathcal{L}_{norm}$.} $\odot$ denotes pixel-wise multiplication and $\oplus$ denotes pixel-wise addition.}
\label{1}
\end{figure}

\begin{figure}[th]
\centering
\setlength{\abovecaptionskip}{0pt}
\includegraphics[width=0.45\textwidth,height=0.25\textheight]{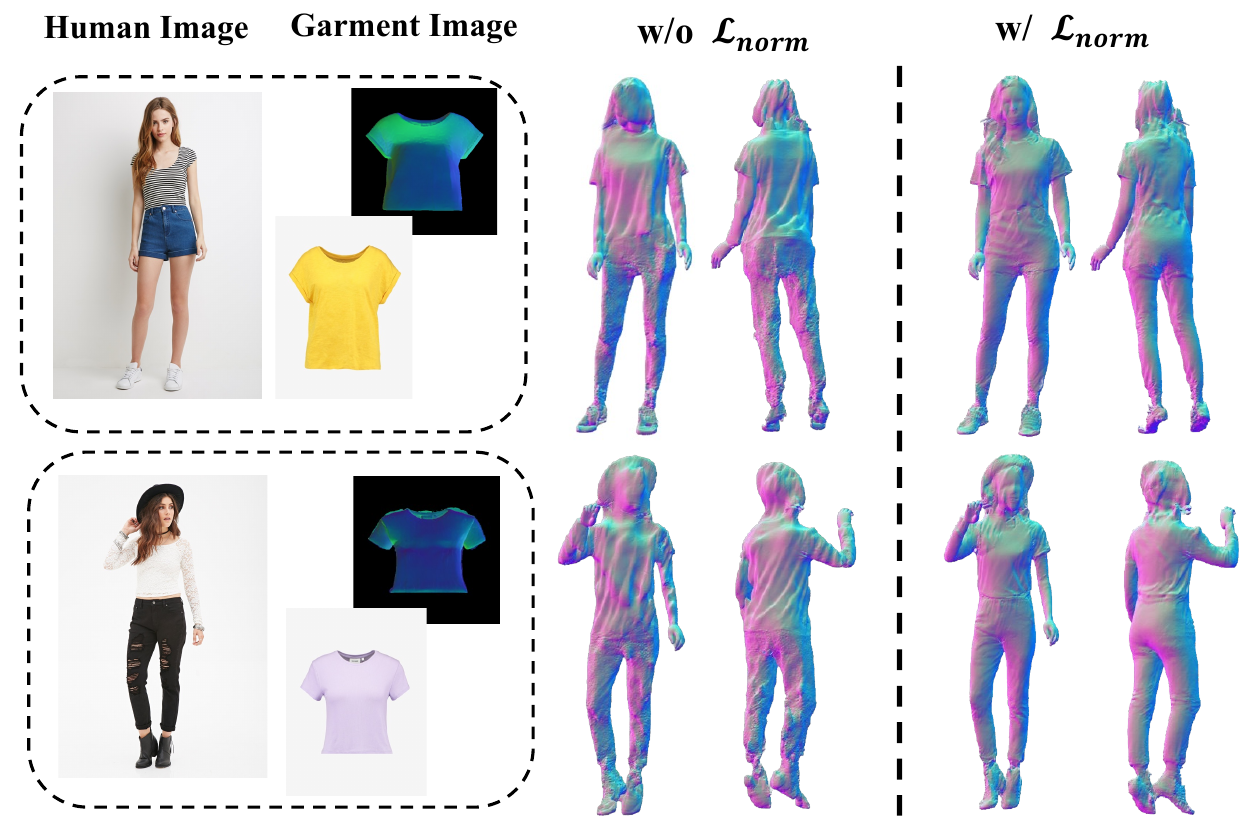}
\caption{\textbf{Ablation study for $\mathcal{L}_{norm}$.}}
\label{2}
\end{figure}

As shown in Fig. \ref{1}, given the source human image $\mathcal{H}_I$, we can obtain its pose map using OpenPose \cite{cao2017realtime} and a partial segmentation map $\hat{m}$ that excludes the try-on regions. With these condition maps and the target text prompts, we can generate the pseudo human image $\mathcal{H}_{I'}$. Using normal map estimation \cite{xiu2022icon}, we can separately obtain the normal maps for $\mathcal{H}_{I'}$ and $\mathcal{H}_I$, denoted as $\mathcal{H}_{I'}^n$ and $\mathcal{H}_I^n$ respectively. We can extract the mask of the garment area, including parts of the arms, from $\mathcal{H}_{I'}$ using SAM \cite{ren2024grounded}, denoted as $m$. Based on $m$ and $\hat{m}$, we extract the corresponding partial normal maps from $\mathcal{H}_{I'}^n$ and $\mathcal{H}_I^n$, respectively. Finally, we obtain a normal map $\mathcal{H}_I^{n'}$ that reflects the geometry with desired garments and source identity. $I^n$ is the rendered normal map, and $\mathcal{L}_{norm}$ is calculated by applying pixel-wise MSE loss between $\mathcal{H}_I^{n'}$ and $I^n$. As shown in Fig. \ref{2}, when the geometry is generated without using the normal loss, the result lacks local details and appears overly smooth.

\begin{figure}[th]
\centering
\setlength{\abovecaptionskip}{0pt}
\includegraphics[width=0.45\textwidth,height=0.15\textheight]{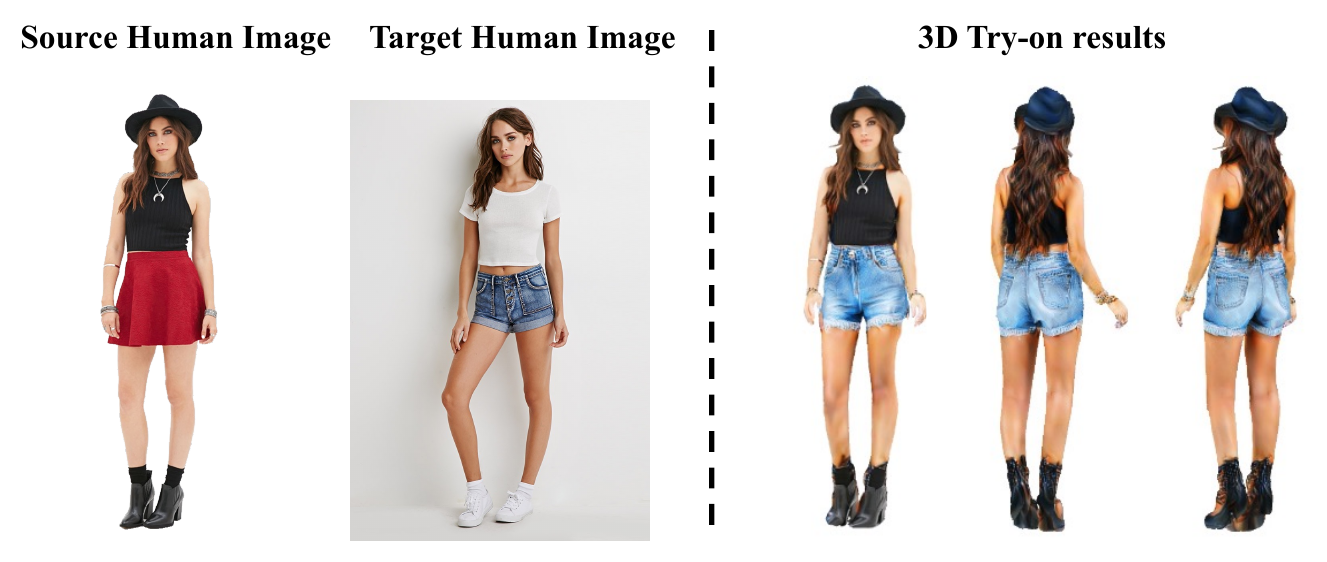}
\caption{\textbf{Template-free Garment Transfer.}}
\label{3}
\end{figure}

\begin{figure}[th]
\centering
\setlength{\abovecaptionskip}{0pt}
\includegraphics[width=0.45\textwidth,height=0.32\textheight]{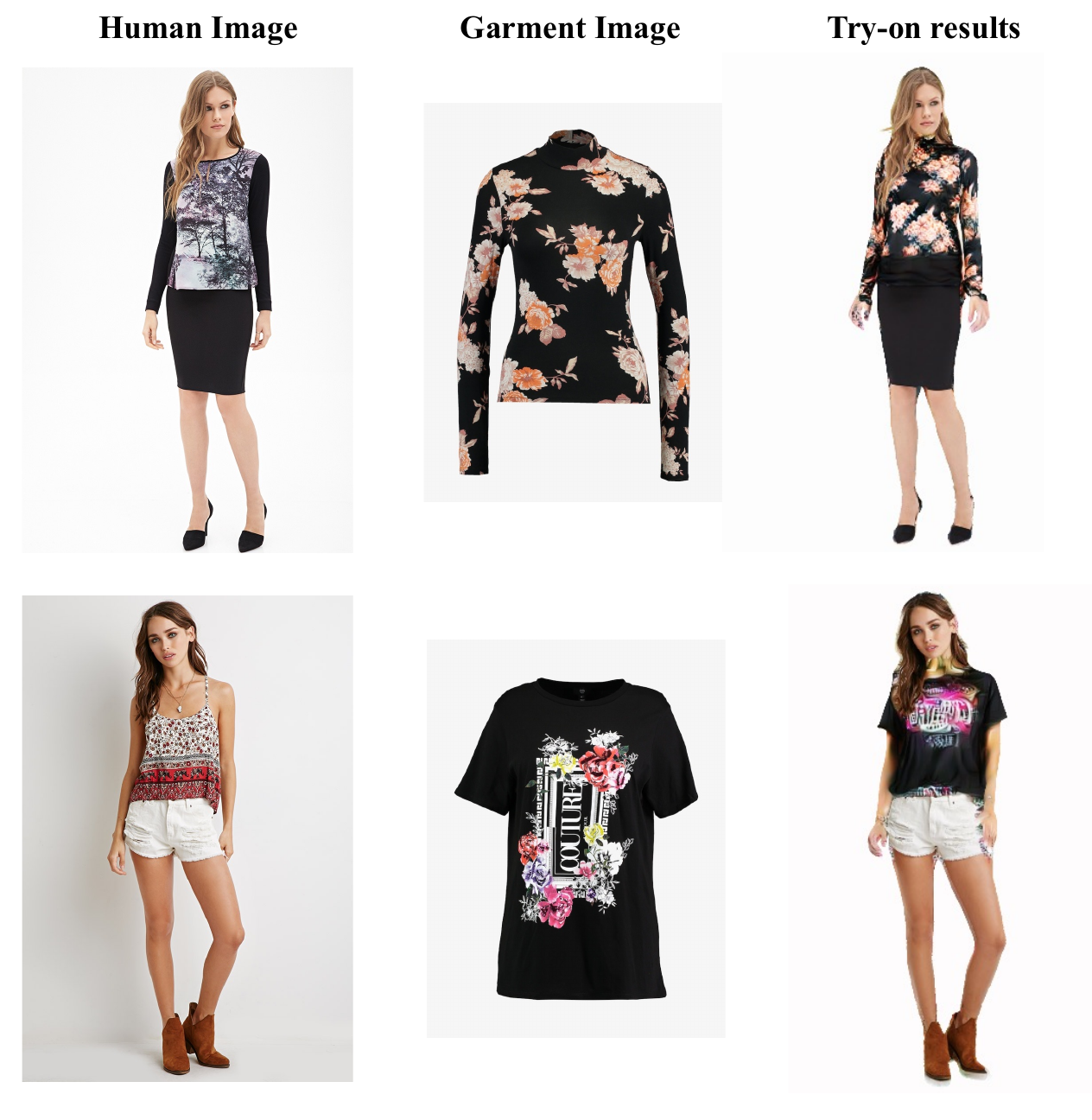}
\caption{\textbf{When the garment image has complex patterns, the generated texture may fail to accurately reproduce these patterns.}}
\label{4}
\end{figure}

\section{Datasets}

Given a human image $\mathcal{H}_I$ and a garment image $\mathcal{H}_g$, we can generate descriptive text prompts for each separately using ChatGPT-4o, denoted as $\mathcal{T}_I$ and $\mathcal{T}_g$. The text prompt $\mathcal{T}_I$ includes descriptions of hair, hat, garments, and shoes. We replace the corresponding parts of $\mathcal{T}_I$ with those from $\mathcal{T}_g$ to generate target text prompts. For example, if $\mathcal{T}_g$ is $\texttt{`}$a black sleeveless crop top$\texttt{`}$ and $\mathcal{T}_I$ is $\texttt{`}$A woman wears a white short-sleeved top and a high-waisted orange pencil skirt. She also wears black ankle boots and has long wavy brown hair.$\texttt{`}$, then the target text prompt would be $\texttt{`}$A woman wears a black sleeveless crop top and a high-waisted orange pencil skirt. She also wears black ankle boots and has long wavy brown hair.$\texttt{`}$.

\section{More Applications}
Our method also supports template-free garment transfer by extracting the target garment directly from the human image. As shown in Fig. \ref{3}, our approach can transfer the shorts from the target human image to the source human image and generate 3D try-on results that can be viewed from any angle.

\section{Limitations}

Unlike traditional 2D image-based try-on methods \cite{kim2024stableviton} that explicitly warp garments and blend them with human images, IPVTON implicitly aligns the image features generated by the IP-Adapter's \cite{ye2023ip} image prompt encoder with 3D objects. However, these image features lack fine-grained details. As shown in Fig. \ref{4}, when the garment images contain complex patterns, IPVTON can generate coarse textures with high-level semantic information of the target garments, but the patterns are not accurately reproduced.

\bibliography{aaai25}